\documentclass[acmsmall,nonacm]{acmart}

\graphicspath{{figures/}}

\usepackage[figuresonly,nomarkers,nofiglist,fighead]{endfloat}

\newenvironment{keyinsights}{%
  \par\medskip\noindent
  \begin{minipage}{0.98\columnwidth}%
  \rule{\linewidth}{1pt}\par\vspace{3pt}%
  {\sffamily\bfseries Key Insights}\par\vspace{2pt}%
  \rule{\linewidth}{0.4pt}\par\vspace{5pt}%
  \small
}{%
  \par\vspace{5pt}\rule{\linewidth}{1pt}%
  \end{minipage}%
  \par\medskip
}

\newenvironment{cacmsidebar}[1]{%
  \par\medskip\noindent
  \begin{minipage}{0.98\columnwidth}%
  \rule{\linewidth}{1pt}\par\vspace{3pt}%
  {\sffamily\bfseries Sidebar: #1}\par\vspace{2pt}%
  \rule{\linewidth}{0.4pt}\par\vspace{5pt}%
  \small
}{%
  \par\vspace{5pt}\rule{\linewidth}{1pt}%
  \end{minipage}%
  \par\medskip
}

\begin{document}

\title{How To Build A Modern Diffusion Language Model}

\author{Volodymyr Kuleshov}
\email{kuleshov@cornell.edu}
\affiliation{%
  \institution{Cornell University}
  \city{New York}
  \state{NY}
  \country{USA}
}

\author{Marianne Arriola}
\email{marriola@cs.cornell.edu}
\affiliation{%
  \institution{Cornell University}
  \city{New York}
  \state{NY}
  \country{USA}
}

\author{Yair Schiff}
\email{yzs2@cornell.edu}
\affiliation{%
  \institution{Cornell University}
  \city{New York}
  \state{NY}
  \country{USA}
}

\author{Gilad Turok}
\email{gt345@cornell.edu}
\affiliation{%
  \institution{Cornell University}
  \city{New York}
  \state{NY}
  \country{USA}
}

\author{Guanghan Wang}
\email{gw348@cornell.edu}
\affiliation{%
  \institution{Cornell University}
  \city{New York}
  \state{NY}
  \country{USA}
}

\begin{abstract}
\end{abstract}

\maketitle

\begin{keyinsights}
\begin{itemize}
\item Discrete diffusion generates language by iteratively refining an initial noisy sequence. 
      This generative process unlocks fast generation at thousands of tokens per second, and supports built-in error correction, guidance, and new forms of inference-time scaling.
  \item Mainstream diffusion models are based on masked or uniform-state diffusion, and
  rely on specialized techniques for iterative refinement, fast sampling, and post-training.
  \item Discrete diffusion scales to billions of parameters when modeling biological sequences and language.
By accelerating generation, diffusion holds potential to improve scaling laws for language modeling.
\end{itemize}
\end{keyinsights}

\section{From Sequential to Parallel Language Generation}
\label{sec:bottleneck}

Large language models (LLMs) have achieved widespread success in recent years,
driven largely by scaling laws for pretraining on large text
corpora~\cite{kaplan2020scaling}. However, mainstream LLMs have fundamental
limitations: they generate text \emph{autoregressively}, producing outputs
sequentially from left to right by predicting one token at a time conditioned on
the tokens generated so far. This sequential decoding process causes several
problems in practice. Most notably, producing $n$ tokens requires calling the
language model sequentially $n$ times, leaving parallel GPU hardware
underutilized during inference. This bottleneck has grown increasingly acute as
improvements in LLM capabilities have shifted toward scaling post-training and
inference-time compute~\cite{snell2025testtime}---such as reinforcement
learning~\cite{zhao2025d1, wang2025d2} and long chains of test-time
reasoning---both of which are gated by token generation speed.

Autoregressive decoding also constrains how a sequence is constructed. Because
tokens are emitted one by one and never revisited, mistakes are permanent: an
early error cannot be corrected later and instead propagates through the rest
of the output. Moreover, because generation proceeds strictly left to right,
each token is chosen using only preceding context, never future context. 

This
article focuses on \emph{discrete diffusion}~\cite{sohldickstein2015deep, austin2021structured}, an alternative paradigm for
language modeling that generates tokens in parallel rather than sequentially,
addressing these limitations and offering a promising path toward accelerating
post-training and inference-time scaling.

\subsection{Discrete Diffusion Models}

\begin{figure}[t]
  \centering
  \includegraphics[width=0.94\columnwidth]{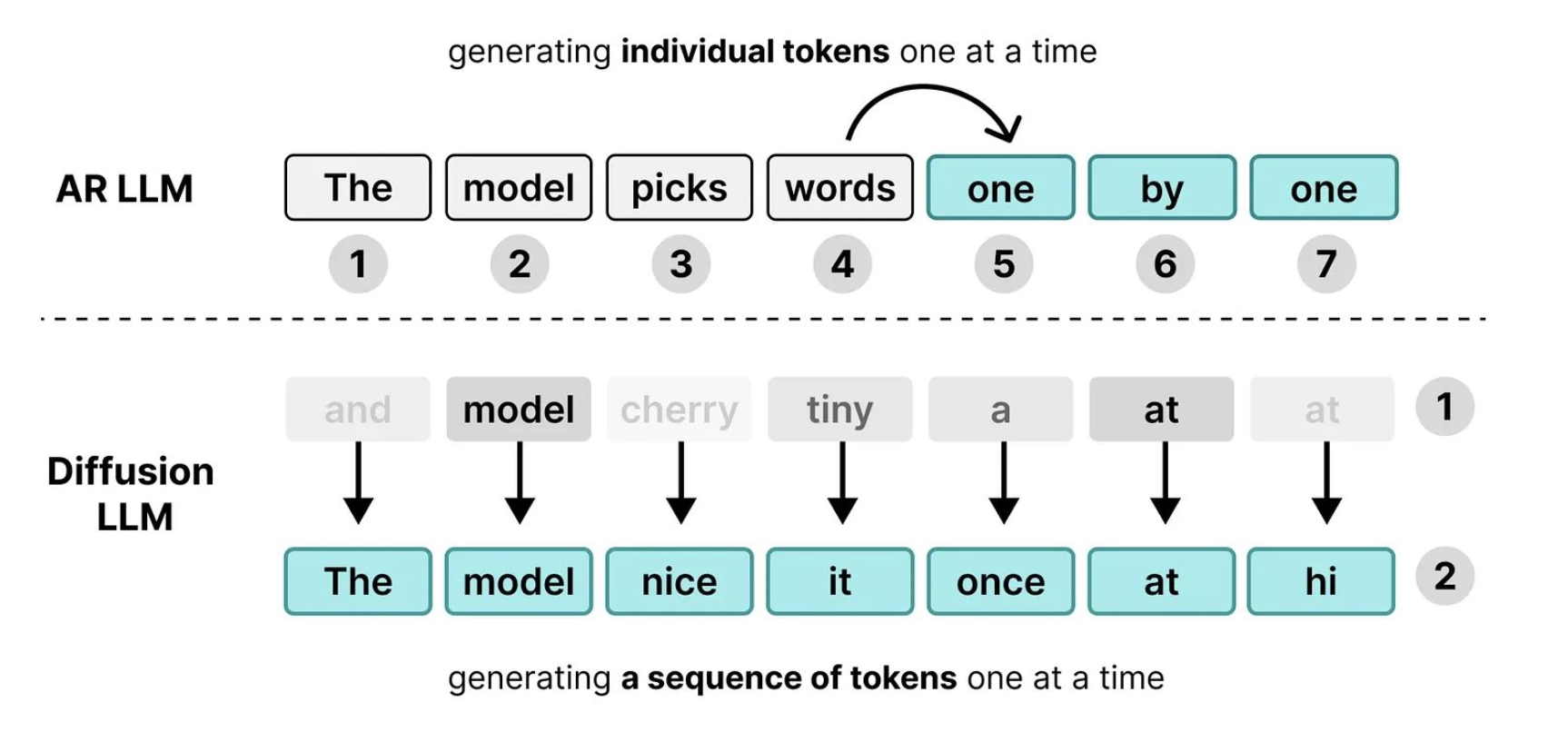}
  \caption{Autoregressive models generate one token per step, left to right, so
  a sequence costs as many forward passes as it has tokens (top). Diffusion
  models start from a full-length masked draft and refine every position in
  parallel over a small number of rounds (bottom). The number of steps is set
  by the refinement schedule rather than by the length of the output.
  Figure credit: Maarten Grootendorst.}
  \label{fig:ar-vs-diffusion}
\end{figure}

Rather than
emitting an output one element at a time, a diffusion model starts from an
initial noisy or masked sequence and iteratively refines the entire sequence in
parallel over a series of denoising steps (Figure~\ref{fig:ar-vs-diffusion}).
Because each step updates multiple tokens simultaneously, the cost of
generation is governed by the number of refinement steps rather than the length
of the output, and each refinement step can produce multiple tokens. This enables generation speeds of thousands of tokens per second on
standard GPUs~\cite{khanna2025mercury, google2026diffusiongemma}. 

Diffusion models can also trade off speed for quality by varying
the number of refinement steps, providing a new axis of inference-time compute
scaling~\cite{wang2025remasking}. Furthermore, because every step attends to bidirectional context and
refines the sequence globally, diffusion can correct earlier mistakes during
generation~\cite{campbell2022continuous, wang2025remasking}---making it particularly effective for controllable generation in scientific domains, such as for designing proteins or regulatory DNA in biology~\cite{hayes2025simulating, boshar2025ntv3}.


While diffusion models quickly became mainstream for generating images,
applying diffusion to discrete data was long an open problem~\cite{austin2021structured, li2022diffusionlm}. A
turning point came in 2024, when discrete diffusion models began to match
autoregressive models in quality at small scale~\cite{lou2024sedd,
sahoo2024simple, shi2024md4}. Over the past two years, discrete diffusion has scaled to
tens and hundreds of billions of parameters across biological sequence
modeling~\cite{hayes2025simulating, boshar2025ntv3} and natural language~\cite{nie2025large}, with
commercial and open-weights releases from leading industry labs including
Mercury~2 from Inception~\cite{khanna2025mercury}, Gemma Diffusion from
Google~\cite{google2026diffusiongemma}, and Nemotron Diffusion from
NVIDIA~\cite{fu2026nemotronlabsdiffusion, reda2026twotower}. 

This article is a survey of key ideas in discrete diffusion, and a high-level tutorial that describes how modern diffusion LLMs are built. We introduce the foundations of discrete diffusion
(Section~\ref{sec:masked}), describe the additional building
blocks of modern diffusion LLMs (Section~\ref{sec:blocks}), survey their
applications in science and language today (Section~\ref{sec:practice}), and
examine how parallel generation may reshape scaling laws for language modeling
(Section~\ref{sec:outlook}).

\section{Understanding Discrete Diffusion}
\label{sec:masked}

We begin with a brief introduction to discrete diffusion, using as an example
a simple framework called masked diffusion.
While simple, it now forms the basis of most open-source diffusion language models.
Later sections introduce the more advanced extensions needed to build such models.

\subsection{Generating by Iterative Denoising}

The central concept underlying diffusion is denoising \cite{sohldickstein2015deep,ho2020ddpm}.
Instead of generating an output in one shot, a diffusion model produces it step by step,
starting from random noise and removing a little of it at every step.
Generating outputs through many small steps is far simpler than producing everything at once,
which is what makes diffusion models so effective.

How does diffusion generate by denoising?
Every diffusion model implements the same blueprint, consisting of two complementary processes \cite{sohldickstein2015deep}.
At training time, a forward process turns each datapoint into pure noise, one step at a time.
A reverse process then learns to retrace this transformation in the opposite direction,
turning pure noise back into a datapoint.
The reverse process is a machine learning model, trained on the data-to-noise trajectories the forward process produces.
At inference time, applying it to new random noise generates new data \cite{ho2020ddpm,song2021sde}

\subsection{Masked Diffusion Language Models}

The main obstacle in bringing diffusion to language is deciding what "noise" should mean for discrete tokens. For images, noise is simply a random Gaussian perturbation of the pixels, but adding continuous Gaussian noise to discrete variables is not well-defined.

A widely used definition of noise for language is masking: replacing ordinary tokens with a special mask token.
The forward process traces a path from clean data to a fully masked sequence by masking a random subset of the tokens at each step.
The reverse process retraces this transformation in the opposite direction,
turning a fully masked sequence back into a datapoint. 

To implement the reverse process, we train a model to fill in the masks in 
the partially masked sequences the forward process produces.
Generation is then simple: starting from a fully masked sequence, we ask the model to infill the masks, we keep a small number of the predicted values, and we repeat. Each pass nudges the sample closer to real data, until an unmasked sequence remains.

Readers who know BERT will recognize the
training procedure~\cite{devlin2019bert}. The main difference is that masked diffusion uses a randomized masking rate, 
which turns BERT into a principled generative models from which we can sample.  
We refer to this framework as a masked diffusion language model (MDLM) \cite{sahoo2024simple}.

\subsection{Autoregressive and Masked Diffusion Language Models: Bridging the Gap}

Formally, an unmasking reverse process defines a latent variable model, in which
partially masked sequences are latent variables.
This latent variable framing makes it possible to compare masked diffusion against autoregressive models using perplexity or log-likelihood, the standard metrics for evaluating language models.
Computing the log-likelihood of a latent variable model is generally intractable, but MDLM admits a strikingly simple formula for its evidence lower bound (ELBO).
\begin{equation}
\label{eq:elbo}
\mathcal{L}_{\mathrm{ELBO}}(x)
  = \mathbb{E}_{t \sim \mathcal{U}[0,1]}\,
    \mathbb{E}_{q(z_t \mid x)}
    \left[ \frac{1}{t}\, \log p_\theta(x \mid z_t) \right].
\end{equation}
Here, $x$ is a datapoint and $t$, drawn uniformly from $[0,1]$, is a masking
rate. We sample a partially masked $z_t$ from the forward process $q(z_t \mid x)$, which masks each token of $x$ independently with probability $t$.
The model $p_\theta$ predicts the clean tokens from that masked input, so the
inner term $\log p_\theta(x \mid z_t)$ is the cross-entropy between its
predictions and the true tokens, which is exactly the BERT loss. The MDLM ELBO averages this loss over all masking rates $t$, and the factor $1/t$ normalizes by the fraction of tokens masked.

\begin{figure}[t]
  \centering
  \includegraphics[width=0.94\columnwidth]{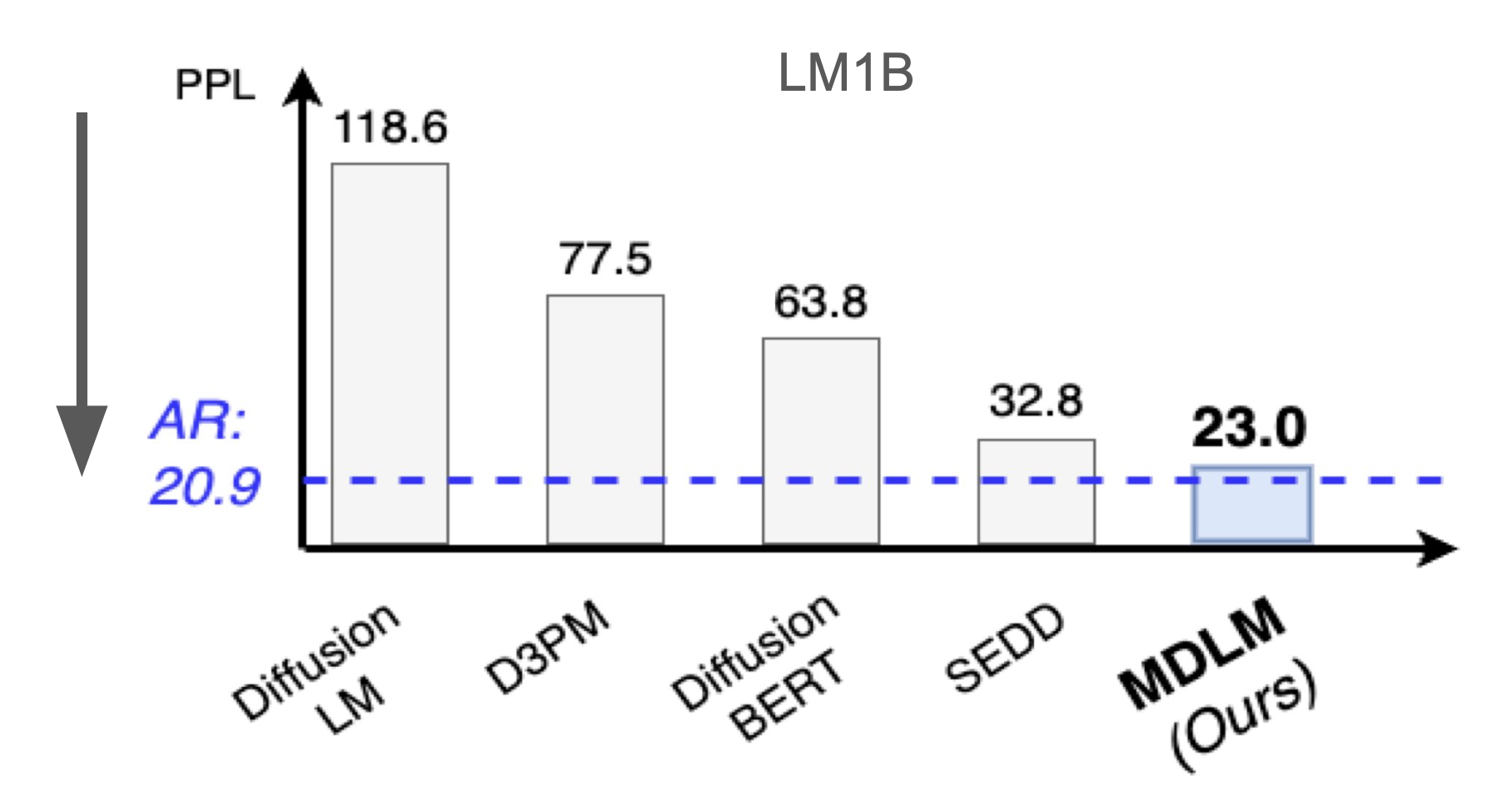}
  \caption{Test perplexity on the One Billion Word benchmark, lower being
  better. Early discrete diffusion models trailed a comparable autoregressive
  transformer by a wide margin. Simplifying the masked diffusion objective closed
  most of the remaining gap, which is what made perplexity comparisons between
  the two families meaningful.}
  \label{fig:ppl}
\end{figure}

For a long time, the two families were separated by a wide margin in quality: earlier discrete diffusion methods trailed
autoregressive
perplexity~\cite{austin2021structured,li2022diffusionlm,lou2024sedd},
and simplified masked diffusion models were among the first to close much of that
gap~\cite{sahoo2024simple,shi2024md4} (Figure~\ref{fig:ppl}).

\begin{figure}[t]
  \centering
  \includegraphics[width=0.94\columnwidth]{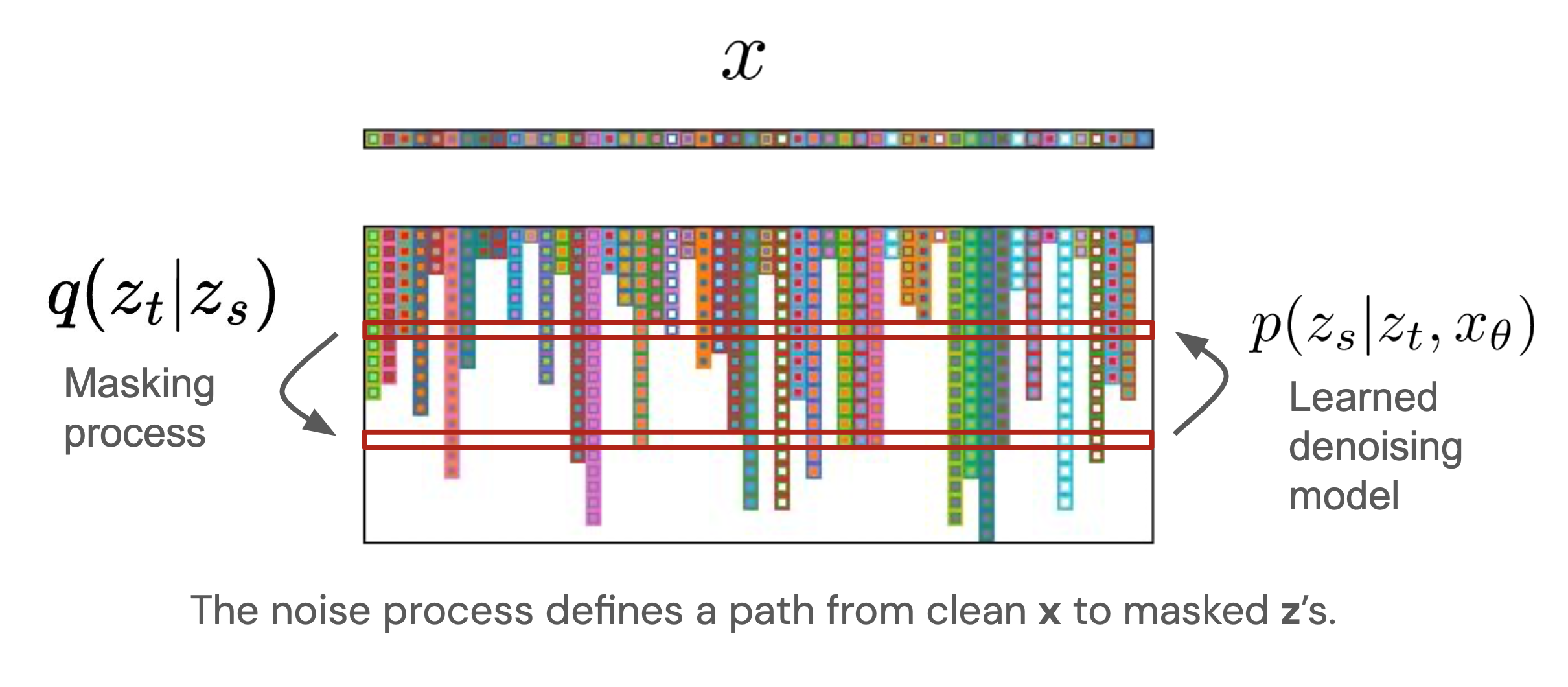}
  \caption{Masked diffusion, forward and reverse. The forward process masks a
  growing fraction of a clean sequence until nothing remains visible, playing the
  role that adding noise plays in an image diffusion model. The reverse process,
  which is the part that is learned, predicts the hidden tokens and so walks the
  same path backwards. Together the two define a latent-variable model over a
  path of increasingly masked sequences. Figure credit: Sasha Rush.}
  \label{fig:masked-diffusion}
\end{figure}

\section{From Simple Masking to Modern Discrete Diffusion Models}
\label{sec:blocks}

Masked diffusion is useful for building intuition, but it does not yet define
a real-world diffusion language model.
This section describes the components that are needed to build real-world models, 
using modern open-weights diffusion LLMs as examples.

\subsection{Efficient Generation of Variable-Length Sequences}

The immediate limitation of MDLM is that generation is only defined for fixed-length sequences, whereas the length of language outputs is variable.
The mainstream solution is block diffusion, which generates blocks of tokens autoregressively and uses diffusion to produce each block~\cite{arriola2025block}.
The block size can vary from dozens to thousands of tokens, and is often chosen to maximize GPU utilization (i.e., so that the arthimetic intensity of the model's forward pass matches that of the underlying hardware).
Gemma Diffusion, for instance, denoises blocks of 256 tokens at a time~\cite{google2026diffusiongemma}.

Block diffusion also restores \emph{KV caching}, the technique that makes
autoregressive generation tractable: once a block is generated, the keys and
values computed in its forward pass can be cached
and reused while generating later blocks. 
%
Other routes to variable-length generation exploit the connection between masked diffusion
and any-order autoregressive models. Set diffusion generalizes block diffusion to
operate over arbitrary sets of positions rather than left-to-right
blocks~\cite{arriola2026setdiffusion}. Edit Flows~\cite{havasi2025editflows} and
FlexMDM~\cite{kim2025flexmdm} instead model generation
as a sequence of insertion, deletion, and substitution operations, which lets the
sequence grow or shrink as it is generated.

\subsection{Iterative Refinement and Inference-Time Scaling}

Diffusion models can naturally correct errors over the course of generation, a process called iterative refinement.
However, standard masked diffusion does not support this \cite{austin2021structured,sahoo2024simple}:
the forward process never unmasks a token that
has already been masked, so the reverse process is never shown an example of revising and improving a
token during generation.

In practice, we must switch to generating from an improved diffusion process, such as remasking~\cite{wang2025remasking}.
Remasking replaces a small subset of previously unmasked tokens with masks, 
allowing them to be later regenerated. 
Remasking is a special case of predictor-corrector methods~\cite{campbell2022continuous}, a general framework for error correction in discrete diffusion.
Other approaches include fine-tuning the denoiser to directly overwrite previously unmasked tokens to correct errors \citep{nie2025large,schiff2026proseco}.

One important feature of iterative refinement is that it
gives discrete diffusion a new form of inference-time compute scaling~\cite{wang2025remasking}: increasing the
number of denoising steps lets the model spend more time refining and correcting
errors, which improves quality (Figure~\ref{fig:scaling}). In autoregressive models, the number of sampling steps
is fixed and inference-time scaling instead comes from chain-of-thought reasoning; adjusting
denoising effort gives diffusion models a novel, orthogonal axis for scaling compute.

\paragraph{Uniform-state diffusion}
Alternatively, we can abandon masking altogether and use a different form of corruption (Figure~\ref{fig:uniform}). In uniform-state diffusion, the most popular alternative to masking,
the forward process replaces tokens in a given sequence with ones sampled uniformly at random from the vocabulary \cite{austin2021structured,schiff2024discreteguidance,sahoo2025duo}. The reverse process begins from a random sequence and flips tokens until the result looks like data.
Error correction comes for free: the reverse process can always revise a token it believes was an erroneous flip.

Masked diffusion models
typically train faster and reach better perplexities, while uniform-state models' error-correction capabilities support faster sampling~\cite{sahoo2025duo} and better controllability~\cite{schiff2024discreteguidance}.
Gemma Diffusion is a
uniform-state model~\cite{google2026diffusiongemma}. Like masked diffusion, uniform-state diffusion admits a
simplified evidence lower bound that improves
training~\cite{schiff2024discreteguidance}. Both uniform and masked diffusion are special cases of the D3PM framework~\cite{austin2021structured} and can be combined via interpolation~\cite{vonrutte2025gidd}.

\subsection{Diffusion Samplers for Ultra-Fast Generation}


Diffusion models can generate sequences much faster than autoregressive models by outputting or refining multiple tokens per step instead of just one, often achieving speeds of thousands of tokens per second \cite{khanna2025mercury,google2026diffusiongemma}.
However, plain masked diffusion is prone to errors in this regime: tokens produced in the same step can be inconsistent (e.g., ``she sell" instead of ``she sells'' or ``they sell").
Fast diffusion-based generation therefore combines a reverse process that supports error correction (e.g., uniform noise or remasking) with a specialized, accelerated sampler \cite{kim2025train,sahoo2025duo}.

Mainstream diffusion samplers rely on the confidence of the model's predictions to guide sampling, e.g., to determine which positions to unmask or to commit to the final output. Common heuristics commit the most confident tokens \cite{kim2025train}, or ones with low entropy \cite{benhamu2025ebsampler}. Such heuristics greatly improve both generation quality and perplexity of standard masked diffusion \citep{turok2026duel}, and are used in every open-source diffusion model \cite{google2026diffusiongemma,nie2025large,fu2026nemotronlabsdiffusion}.

Further improving generation speed relies on acceleration techniques derived from progressive
distillation~\cite{salimans2022progressive}: a model is trained on its own
generations to skip a step, and the procedure is applied recursively, halving the
number of sampling steps each round. In discrete diffusion models the idea
generalizes to self-distillation through time~\cite{deschenaux2024sdtt} and
discrete consistency distillation~\cite{sahoo2025duo}.

Progressive distillation can be interpreted as a form of on-policy training. Standard
diffusion training is off-policy: the reverse process is trained on samples drawn
from the forward process, not on the samples it will actually produce at generation
time. Training on the model's own samples closes that gap, and doing so as a
post-training step---rather than during primary training, where it would be too
expensive---is what makes it tractable \cite{google2026diffusiongemma}.

\subsection{Controllable Generation}

The goal of controllable generation is to produce outputs with a desired property, e.g., a molecule with high binding affinity to a target site \cite{schiff2024discreteguidance}.
The ability to make iterative global changes (rather than committing
to irreversible local edits) makes diffusion particularly effective at this task \cite{li2022diffusionlm}.

\begin{figure}[t]
  \centering
  \includegraphics[width=0.94\columnwidth]{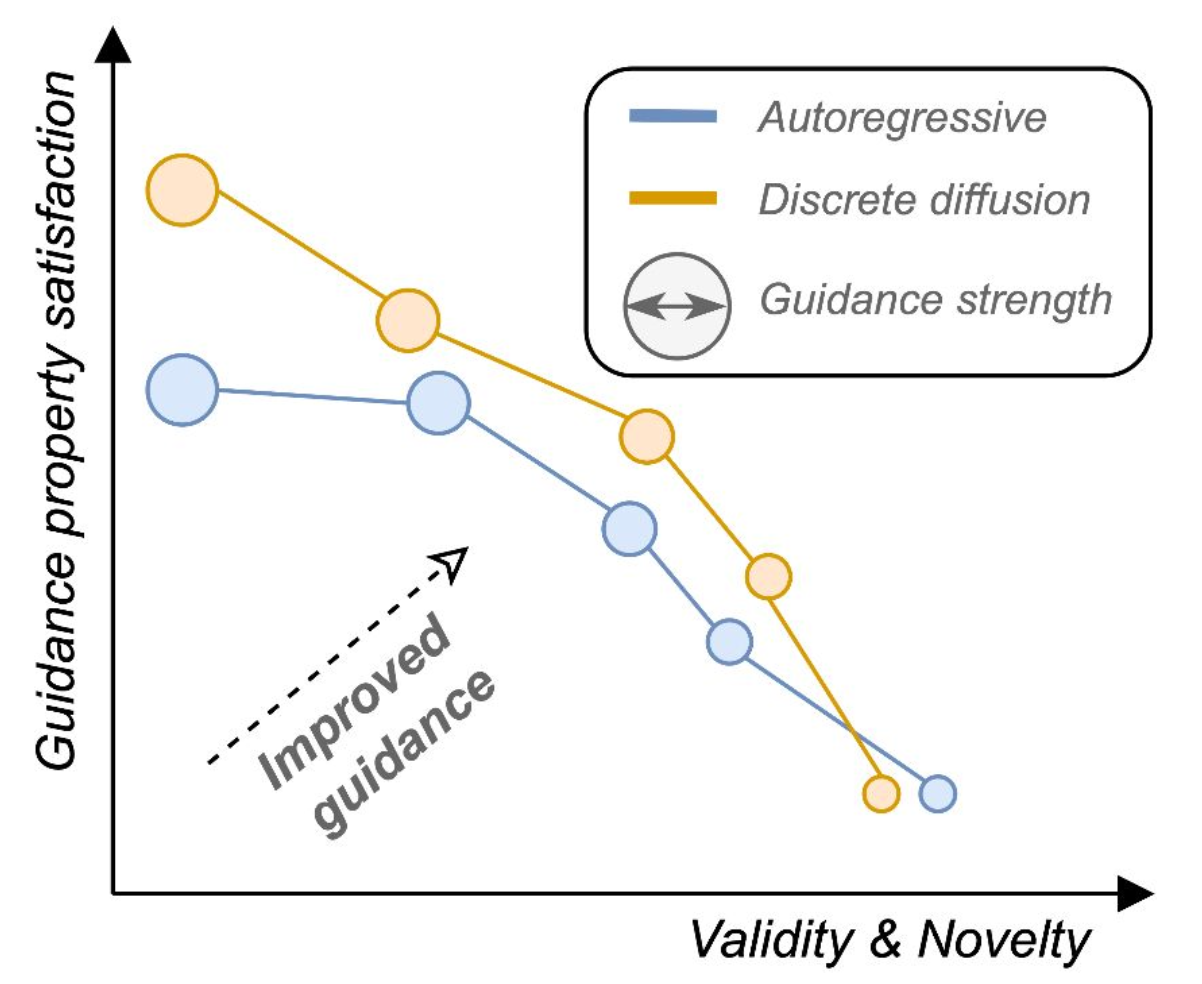}
  \caption{Controllable generation as a trade-off between how natural a sample
  looks and how well it satisfies a target property, traced out by varying the
  guidance strength. Discrete diffusion attains a better frontier than
  autoregressive generation: at a comparable level of naturalness it reaches
  higher property satisfaction.}
  \label{fig:pareto}
\end{figure}

In practice, controllability requires navigating a trade-off between 
producing natural-looking outputs and ones satisfying the desired property \cite{schiff2024discreteguidance}. 
For example, a model could produce
molecules that look natural but lack the binding affinity being sought;
optimizing hard for binding affinity instead can yield molecules that are neither
natural-looking nor synthesizable. Discrete diffusion tends to strike a better trade-off between the two (Figure~\ref{fig:pareto}).

Classifier-based and classifier-free guidance are two popular approaches to controllable generation for diffusion models \cite{dhariwal2021guidance,ho2022cfg}.
In classifier-based guidance \cite{dhariwal2021guidance},
given a predictor $p(y \mid x)$ of the target property $y$ from a sample $x$, that predictor is used at
each step to steer generation, provably yielding a sample from the conditional distribution $p(x \mid y) \propto p(y \mid x) p(x)$. Classifier-free guidance \cite{ho2022cfg} dispenses with the separate
predictor and combines a trained conditional model $p(x \mid y)$ and an unconditional model $p(x)$ to sample from a conditional distribution $p_\gamma(x \mid y) \propto p(x \mid y)^\gamma p(x)^{1-\gamma}$, where $\gamma \in \mathbb{R}$ controls the guidance strength.
Both types of guidance extend naturally to discrete
diffusion~\cite{schiff2024discreteguidance}, and are especially effective when combined with uniform noise or remasking, since both support revising tokens repeatedly as guidance is applied. 
The accompanying sidebar gives the derivations.

\subsection{Architectures for Discrete Diffusion Models}

So far, the algorithms we described have been independent of architecture considerations. For example, in masking diffusion, we fill-in masks using a transformer \cite{devlin2019bert,austin2021structured} when modeling text and a state-space model when modeling DNA \cite{sahoo2024simple}.
Diffusion specifies techniques for training and sampling from a neural network, not its architecture. 

Nonetheless, certain architectural elements play an important role in discrete diffusion. Standard masked diffusion models (e.g., MDLM, LLADA \cite{sahoo2024simple,nie2025large}) are effectively encoder-only, like BERT, in
contrast to the decoder-only architecture of autoregressive models like GPT. That
choice has a cost: an encoder-only model must invoke the full network at every
denoising step, which is computationally expensive when there are many steps.
Modern diffusion models (e.g., Gemma, Nemotron \cite{google2026diffusiongemma,reda2026twotower}) instead rely on 
an encoder--decoder architecture \cite{arriola2025e2d2}: an encoder
computes a representation of the tokens generated so far, and a lightweight decoder iteratively
refines the noised sequence conditioned on the encoder's output. 
The per-step iterative denoising cost greatly improves with a lightweight decoder. 

Another important architectural element in modern diffusion models (e.g., Gemma Diffusion \cite{google2026diffusiongemma}) are residual connections across denoising steps, a process also known as self-conditioning \cite{chen2023analogbits} or loopholing \cite{jo2026loopholing}. The discrete sequence generated at each step of denoising lacks the full information in the hidden states of the denoiser. Passing these embeddings from one step of denoising to another via a skip connection greatly improves quality and sampling speed \cite{google2026diffusiongemma}.



\subsection{Training and Post-Training Discrete Diffusion Models}

Language models are developed using a series of pre-training and post-training steps, and the same recipe applies in modern diffusion LLMs. While some diffusion LLMs are reported to be pre-trained from scratch (e.g., LLaDA \cite{nie2025large}), others were produced by continuing training from the weights of an autoregressive LLM (e.g.,
Diffusion Gemma~\cite{google2026diffusiongemma}). 

Pre-training from scratch is governed by scaling laws similar to autoregressive pre-training: minimizing perplexity on a large corpus correlates with downstream performance, including benchmarks \cite{nie2025large}. While some scaling laws report that diffusion requires 16x more compute to reach the same perplexity, later studies further improved pre-training efficiency, and it remains an active research area \cite{wang2026computegap}. Repurposing autoregressive models into diffusion can be done by fine-tuning the models to do unmasking \cite{fu2026nemotronlabsdiffusion}, or by attaching a decoder head to an autoregressive encoder \cite{arriola2025e2d2,reda2026twotower,google2026diffusiongemma}.


Post-training usually involves standard supervised fine-tuning (SFT) with reinforcement learning (RL) stages \cite{zhao2025d1, wang2025d2}.
While the SFT stage optimizes the diffusion loss over SFT data, the RL stage uses policy gradient methods to optimize a reward function in an environment. Policy gradient methods need the likelihood of a sampled trajectory, which is intractable for diffusion models. 
One solution is diffu-GRPO, a
critic-free policy gradient algorithm that estimates trajectory log-probabilities
with a mean-field approximation. 
Alternatively, models that support any-order decoding can use an exact one-pass estimator that generalizes to an approximate estimator with a tunable compute-accuracy trade-off for general models \citep{wang2025d2}.


\begin{cacmsidebar}{Discrete Guidance, Derived}
\emph{Classifier-based guidance.} Any conditional reverse process factors, by
Bayes' rule, into a predictive term and an unconditional term. Given a classifier
$p(y \mid z_t)$ of the property $y$ from a noisy sequence and an unconditional
diffusion model, the two can be recombined into a conditional model,
\[
  \log p^{(\gamma)}(z_s \mid z_t, y)
    = \gamma \log p(y \mid z_s, z_t) + \log p(z_s \mid z_t),
\]
where $\gamma > 0$ is a guidance strength trading the target property against the
model's own preferences. This is discrete classifier-based guidance. For a single
token the right-hand side is easy to normalize, since one simply sums over the
vocabulary; extending it to a full sequence requires further normalization
machinery~\cite{schiff2024discreteguidance}.

\medskip
\emph{Classifier-free guidance.} This removes the separate classifier. Applying
Bayes' rule to the predictive term above and absorbing everything independent of
$z_s$ into the normalizing constant leaves a combination of a conditional and an
unconditional reverse model,
\[
  \log p^{(\gamma)}(z_s \mid z_t, y)
    = \gamma \log p(z_s \mid z_t, y) + (1 - \gamma) \log p(z_s \mid z_t).
\]
In practice both distributions are parameterized by the \emph{same} network,
trained by randomly dropping the conditioning signal so that it learns both modes
at once.
\end{cacmsidebar}

\begin{figure}[t]
  \centering
  \includegraphics[width=0.75\columnwidth]{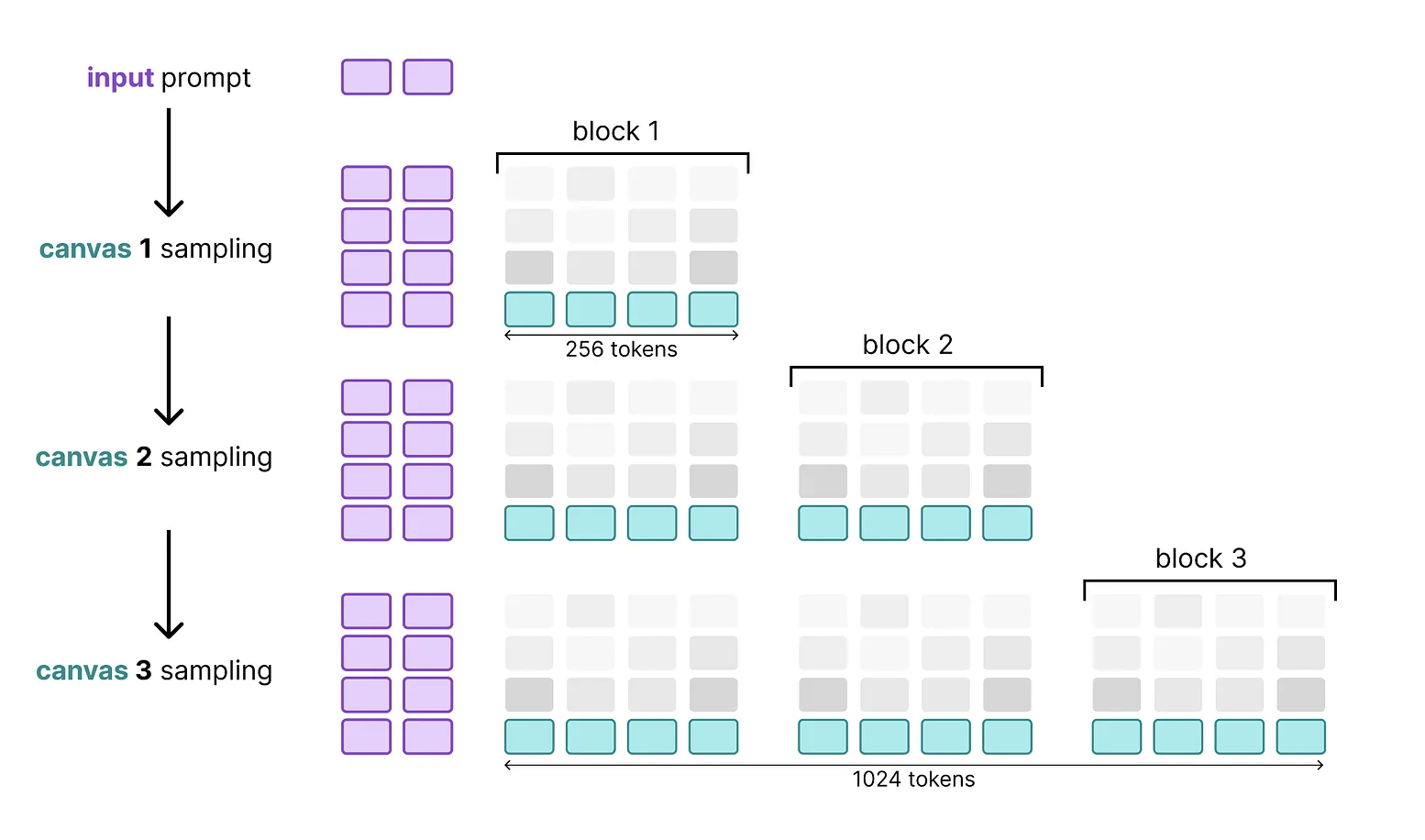}\\[4pt]
  \includegraphics[width=0.6\columnwidth]{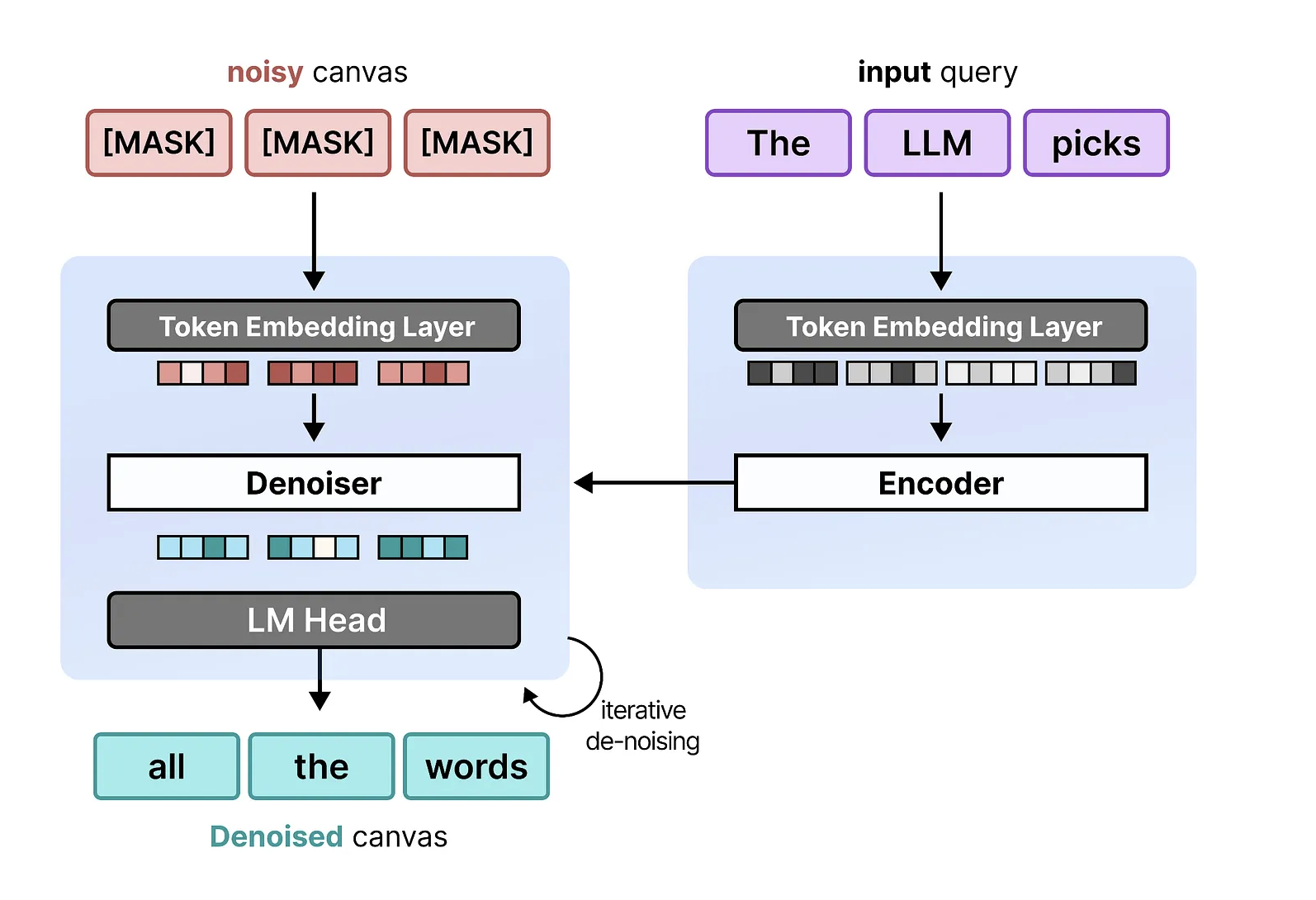}
  \caption{Two extensions that production models rely on. Above, block diffusion
  generates one block of tokens at a time, appending blocks left to right and
  caching the keys and values of earlier blocks much as an autoregressive model
  caches its prefix. Below, the encoder--decoder split assigns a heavy encoder to
  represent the clean context once and a light decoder to denoise the masked
  canvas at every step. Figure credit: Maarten Grootendorst.}
  \label{fig:blocks}
\end{figure}

\begin{figure}[t]
  \centering
  \includegraphics[width=0.94\columnwidth]{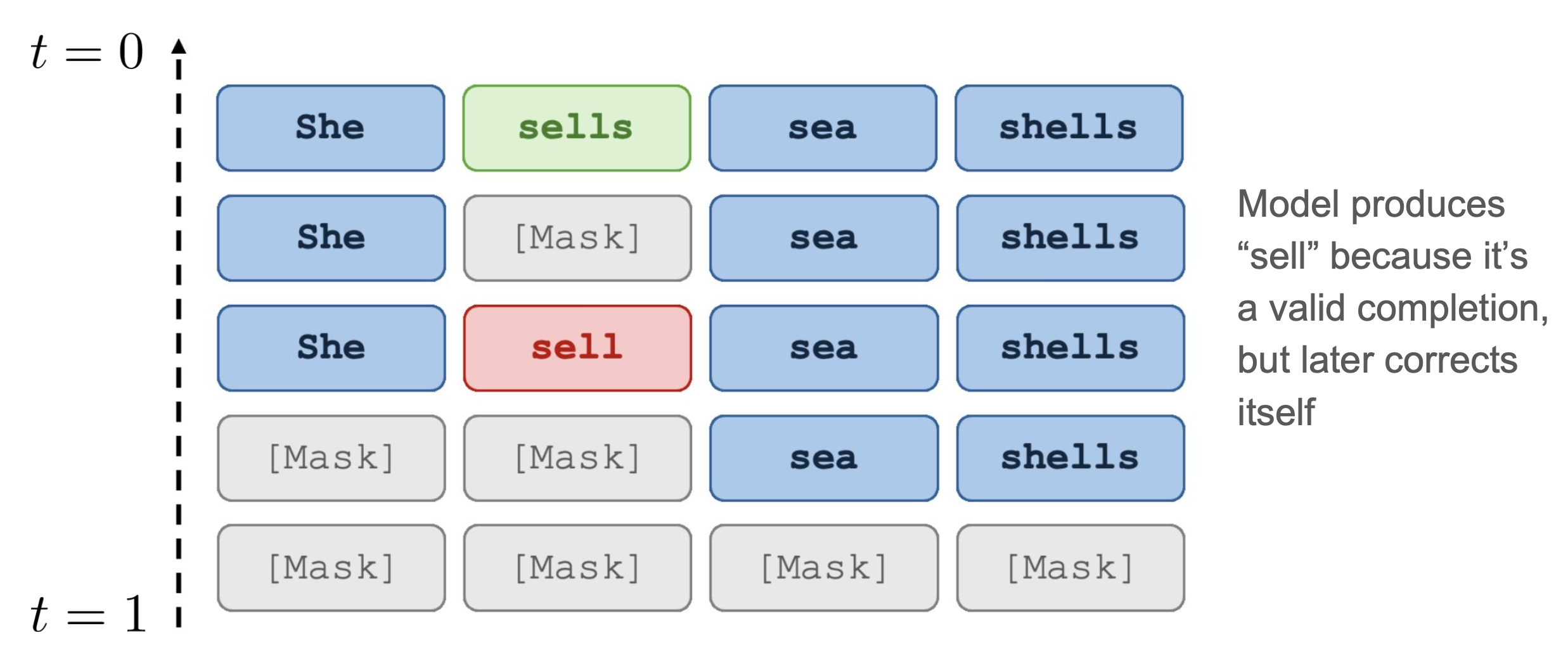}
  \caption{Remasking enables error correction. Reading from a fully masked
  sequence toward a clean one, the model first commits an ungrammatical token
  because it is a locally plausible completion. A later remasking step returns
  that position to the masked state, and with more surrounding context now
  available the model corrects it. Plain masked diffusion cannot do this, because
  its forward process never re-masks and so never teaches revision.}
  \label{fig:remasking}
\end{figure}

\begin{figure}[t]
  \centering
  \includegraphics[width=0.94\columnwidth]{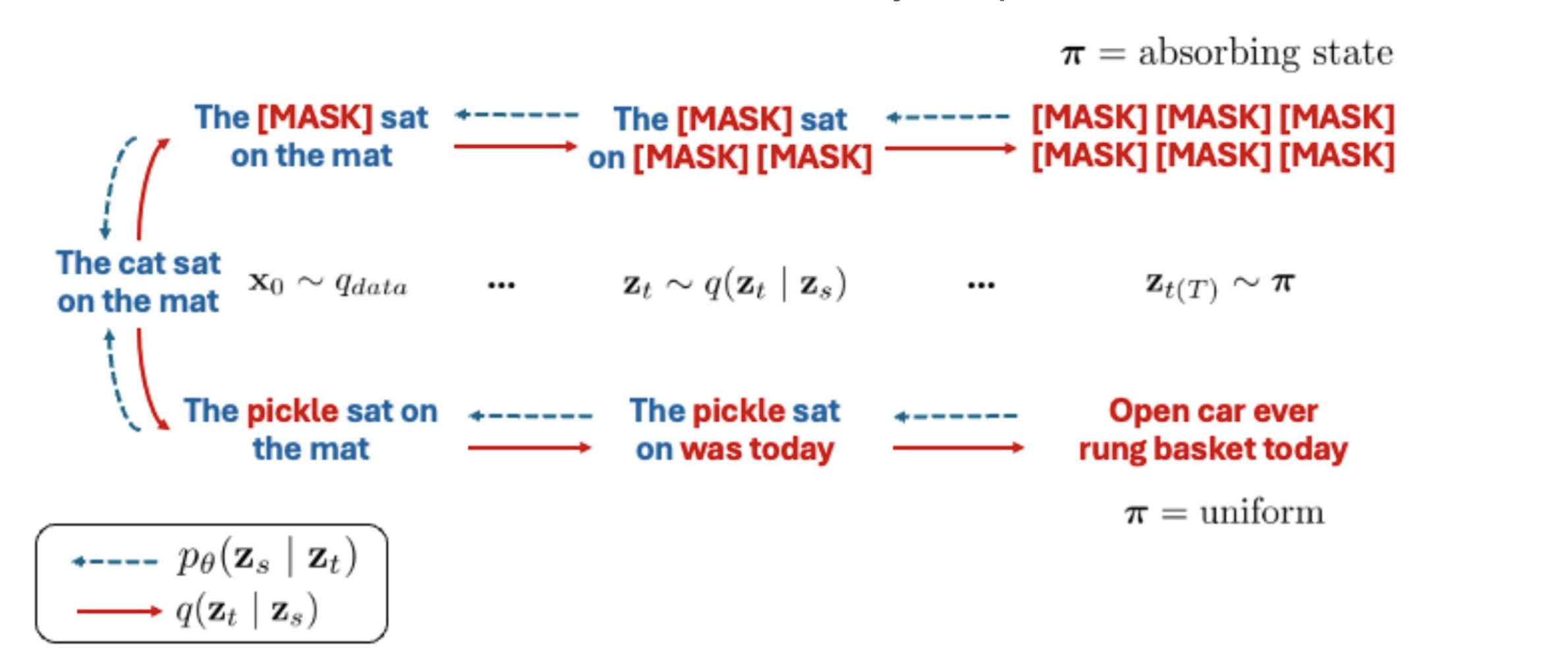}
  \caption{The two discrete corruption processes, applied to the same sentence.
  Masked diffusion progressively replaces tokens with a mask symbol, so a
  position is either settled or blank. Uniform-state diffusion replaces them with
  random vocabulary tokens instead, so intermediate sequences contain no masks and
  every position remains revisable at every step. This is the second route to
  error correction, and an alternative to the remasking shown in
  Figure~\ref{fig:remasking} rather than a successor to it.}
  \label{fig:uniform}
\end{figure}

\begin{figure}[t]
  \centering
  \includegraphics[width=0.94\columnwidth]{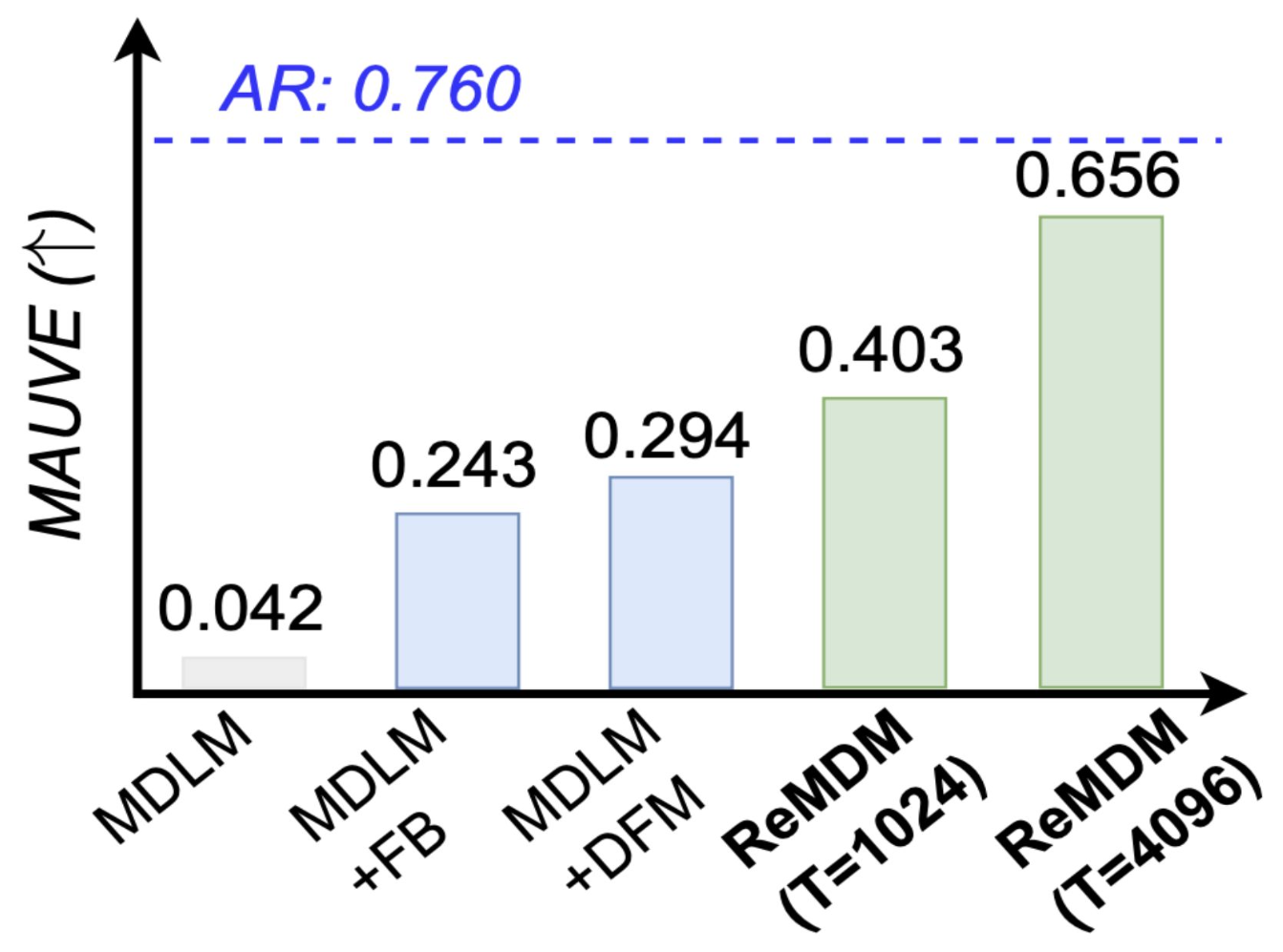}
  \caption{Sample quality against sampling compute, measured by MAUVE, where
  higher is better. Remasking diffusion improves from 0.40 to 0.66 as the number
  of sampling steps rises from 1,024 to 4,096, closing much of the gap to
  autoregressive generation at 0.76 and far exceeding plain masked diffusion at
  0.04. Spending more compute at inference time buys more quality, which is the
  property a scaling law requires.}
  \label{fig:scaling}
\end{figure}

\section{Diffusion Language Models in the Wild}
\label{sec:practice}

Over the past few years, diffusion language models have been scaled to billions of
parameters. This section highlights examples of where they are used today in science
and in language modeling, and describes each model as a combination of the building blocks
from Section~\ref{sec:blocks}.

\subsection{Biological Sequence Modeling}

Some of the earliest successes of diffusion are in scientific applications, particularly
biological sequences. Biological sequences are less likely to exhibit the left-to-right
bias that autoregressive models bake in; scientific applications also frequently need controllable generation, which
is a natural strength of diffusion.

\paragraph{Protein sequences}
The first large-scale application of discrete diffusion was in protein modeling.
ESM3 implements masked diffusion at up to 100 billion parameters, trained on
large amino-acid databases~\cite{hayes2025simulating}. Both its training and its
sampling algorithms match the masked diffusion approach of
Section~\ref{sec:masked}, applied across three interleaved tracks covering
sequence, structure, and function. 

\paragraph{Nucleotide sequences}
Protein activity is heavily regulated
by non-coding genomic sequence. 
DNA language models such as Nucleotide Transformer extend the approach to genomic sequences,
scaling masked diffusion to billions of parameters and more than a trillion tokens of DNA~\cite{boshar2025ntv3}.
Nucleotide Transformer also supports discrete classifier-free guidance combined
with remasking to generate regulatory DNA with specific properties, such as promoting the expression of a gene.


\subsection{Diffusion Large Language Models}

The past year has seen multiple open-weights diffusion language models appear in quick succession.
Table~\ref{tab:models} summarizes how they are assembled from
previously described building blocks.

\paragraph{LLaDA}
LLaDA scaled the masked diffusion approach in Section~\ref{sec:masked} to 8 billion parameters,
combining a masked diffusion backbone with block diffusion at sampling time,
and reporting scaling behavior competitive with autoregressive
models~\cite{nie2025large}. Being open, LLaDA has become the default target for 
research on infernece-time scaling, post-training, multi-modal generation, significantly accelrating the field's progress.

\paragraph{Mercury}
Mercury was the first commercial diffusion LLM. 
Leveraging parallel generation capabilities of diffusion, its achieves speeds of >1000 tokens/second/user on standard GPUs while
matching the quality of speed-optimized frontier models~\cite{khanna2025mercury}.
Throughput at this level was previously reachable only with specialized
inference chips purpose-built to accelerate autoregressive decoding.
Diffusion reaches
comparable speeds on commodity GPUs by changing the algorithm to fit the hardware
rather than building hardware to fit the algorithm. 

\paragraph{Gemma Diffusion}
Gemma Diffusion is an open-weights model from Google~\cite{google2026diffusiongemma}. It combines a uniform-state backbone with block
diffusion for variable-length generation, an encoder--decoder architecture, and
accelerated sampling techniques. This model 
denoises an entire block of tokens per forward pass instead of emitting one token, thus
shifting the generation bottleneck from memory bandwidth to compute.

\paragraph{Nemotron Diffusion}
Nemotron Diffusion is a family of open-weights models from NVIDIA trained with a
joint autoregressive and diffusion objective, implementing masked diffusion and
block diffusion, with recent models adding an encoder--decoder architecture and
scaling to 35 billion parameters~\cite{fu2026nemotronlabsdiffusion}. They report
roughly two to eight times the throughput of comparable autoregressive models
while remaining comparable on quality. A single checkpoint can
fall back to plain autoregressive decoding, allowing a principled comparison of the two generation paradigms.

\begin{table}[t]
  \caption{How three open-weights diffusion language models are built from
  the building blocks of Section~\ref{sec:blocks}.}
  \label{tab:models}
  \small
  \begin{tabular}{@{}llll@{}}
    \toprule
     Model family & \textbf{LLaDA} \cite{nie2025large} & \textbf{Gemma Diffusion} \cite{google2026diffusiongemma} & \textbf{Nemotron Diffusion} \cite{fu2026nemotronlabsdiffusion,reda2026twotower} \\
    \midrule
    Diffusion process   & masking      & uniform-state & masking \\
    Block diffusion size      & 32 & 256           & 16 \\
    Architecture     & decoder-only          & encoder-decoder           & encoder-decoder \\
    Default sampler & max confidence          & entropy bound           & max confidence \\
    Iterative refinement     & remasking, self-correction  & built-in (uniform)      & remasking, self-correction \\
    Training     & from scratch  & from Gemma 26B-A4B      & from Mistral, Nemotron \\
    Parameter count     & 8-100B          & 26B           & 3-35B \\
    \bottomrule
  \end{tabular}
\end{table}


\begin{figure}[t]
  \centering
  \includegraphics[width=0.94\columnwidth]{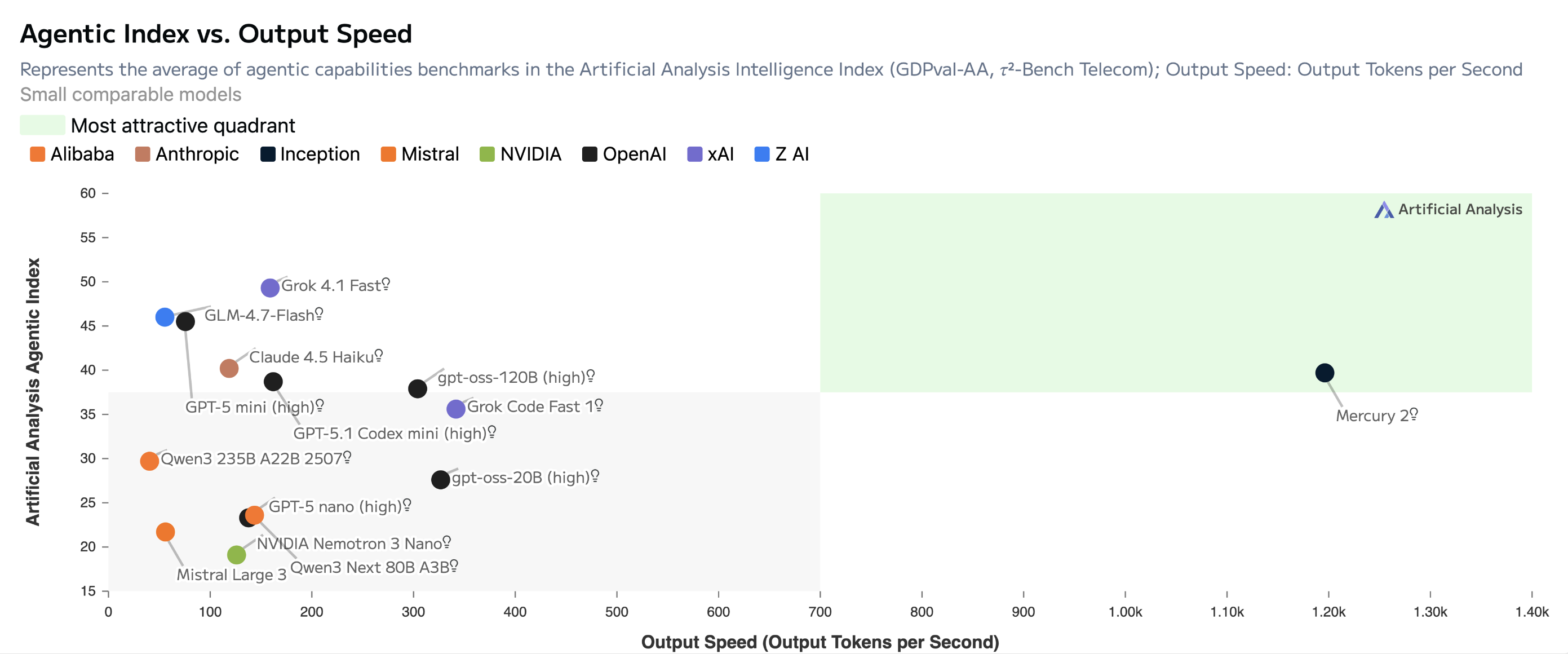}
  \caption{Model quality against output speed for comparable systems, where
  quality is an aggregate agentic benchmark index. The diffusion model sits far to
  the right at roughly 1,200 tokens per second, several times faster than
  autoregressive models of similar quality. Figure credit: Inception.}
  \label{fig:mercury}
\end{figure}

\section{Outlook}
\label{sec:outlook}

In just two years, discrete diffusion models have progressed from small-scale
research prototypes to billion-parameter open-weights and commercial systems
across language and biology. As this article has traced, generating tokens in
parallel unlocks multi-fold speedups in generation
throughput, improved controllability toward target properties, and new
forms of inference-time compute scaling.


A central question is whether diffusion can eventually yield fundamental
improvements over autoregressive models in raw intelligence, rather than merely
making models run faster. 
Recent gains in LLM capabilities have increasingly come from
scaling inference-time compute and post-training~\cite{snell2025testtime}, both
of which are paid for in generated tokens. At inference time, faster generation
lets a model think through longer reasoning chains or explore more candidate
solutions within a fixed latency budget. During post-training, faster
generation enables reinforcement learning over a much larger volume of sampled
rollouts, allowing models to ingest more data and compute in the same
wall-clock time. Diffusion also provides an orthogonal axis of inference-time
scaling that autoregressive decoding lacks: quality can be improved by
denoising a sequence for more steps, rather than only by writing more text
(Figure~\ref{fig:scaling}).

More broadly, history shows that the most successful machine learning
algorithms are those that best utilize the underlying hardware. From 2019 to
2024, pretraining scaling laws were unlocked when the
transformer~\cite{vaswani2017attention} replaced sequential recurrent networks
with a fully parallel training pass that could saturate modern GPUs. If history
is any guide, diffusion models are similarly poised to win the ``hardware
lottery'' for inference---playing the same role for post-training and
inference-time scaling laws that the transformer played for pretraining.

\subsection{Limitations and Open Problems}

Discrete diffusion remains an active research area with both rapid progress and
open problems. We conclude with several active research directions in the field.

\emph{Which noise process is right.} Masked diffusion reaches better
perplexities, while uniform-state diffusion samples faster and is easier to
guide. 
An emerging third option is to map discrete tokens in a continuous space
and run classical forms of diffusion in that space~\cite{li2022diffusionlm, lee2026flow, yang2026replaid}.
These formulations make it easier to
borrow techniques from image generation, such as accelerated few-step samplers,
although the scaling of continuous text diffusion models 
to larger model sizes remains an active research area.

\emph{Faster sampling and post-training.}
Predicting many tokens simultaneously creates a trade-off between speed and
quality. 
Oftentimes, the diffusion model 
has the ability
to provide a quality answer at slow generation speeds (suggesting it knows the answer in its weights), 
but struggles when generating tokens quickly \cite{fu2026nemotronlabsdiffusion}.
This loss of quality also affects reinforcement learning, which also faces the 
challenge of estimating trajectory
likelihoods across unmasking orders~\cite{zhao2025d1, wang2025d2}. 
Finding the
right pairing of noise process, sampler, and post-training objective 
to retain both speed and quality thus remains an open problem.

\emph{Multi-modal generation.}
Because diffusion is already the standard approach for generating continuous
data such as images, video, and audio, discrete diffusion offers a natural path
toward unified multi-modal models built on a single generative paradigm. Early
examples such as ESM3~\cite{hayes2025simulating} already jointly denoise
interleaved tracks of protein sequence, structure, and function, and recent work
has extended language diffusion models~\cite{nie2025large} to vision and
language. How best to combine discrete and continuous modalities---whether by
tokenizing all inputs into a shared space or coupling discrete and
continuous diffusion processes within one model---remains an open problem.

\emph{Co-designing models, systems, and hardware.}
Diffusion models are ideally deployed so that the block size and architecture
match the arithmetic intensity of the hardware they run on. This co-design
could run in both directions, pairing specialized accelerators with custom
kernels and inference stacks tailored to parallel iterative refinement.

Lastly, discrete diffusion has not yet been scaled to the same parameters,
compute, and data as frontier autoregressive models. Scaling laws for diffusion 
are an active research area, and recent studies have further improved
pretraining efficiency~\cite{wang2026computegap}. Closing the scaling gap is a
major focus of ongoing work. It is early to say how fast diffusion research will
progress. The promise, however, is large.

\label{sec:endofbody}

\bibliographystyle{ACM-Reference-Format}
\bibliography{refs}

\end{document}